\documentclass[11pt]{article}

\usepackage[]{acl}

\usepackage{times}
\usepackage{latexsym}
\usepackage{hyperref}
\usepackage{xcolor}
\usepackage{mathrsfs}
\usepackage{amsmath}
\usepackage{multirow}
\usepackage{hyperref}
\usepackage{graphicx}

\usepackage[T1]{fontenc}

\usepackage[utf8]{inputenc}

\usepackage{microtype}

\title{Pre-trained Token-replaced Detection Model as Few-shot Learner}

\author{Zicheng Li \qquad Shoushan Li\thanks{*Corresponding author} \qquad Guodong Zhou \\
Natural Language Processing Lab, Soochow University, China \\
\texttt{20205227019@stu.suda.edu.cn} \\
\texttt{\{lishoushan, gdzhou\}@suda.edu.cn} \\ 
}

\begin{document}
\maketitle
\begin{abstract}
  Pre-trained masked language models have demonstrated remarkable ability as few-shot learners. In this paper, as an alternative, we propose a novel approach to few-shot learning with pre-trained token-replaced detection models like ELECTRA. In this approach, we reformulate a classification or a regression task as a token-replaced detection problem. Specifically, we first define a template and label description words for each task and put them into the input to form a natural language prompt. Then, we employ the pre-trained token-replaced detection model to predict which label description word is the most original (i.e., least replaced) among all label description words in the prompt. A systematic evaluation on 16 datasets demonstrates that our approach outperforms few-shot learners with pre-trained masked language models in both one-sentence and two-sentence learning tasks. \footnote{Our code is available at \url{https://github.com/cjfarmer/TRD_FSL}}
  \end{abstract}
  
  \section{Introduction}
  Few-shot learning aims to learn models with a few examples and the learned models generalize well from very limited examples like humans. Recently, few-shot learning has become an important and interesting research field of intelligence \citep{lake2015human, yogatama2019learning}. Compared to data-rich supervised learning, few-shot learning greatly overcomes the expensive data annotation challenge in reality.
  
\begin{figure*}
    \centering
    \includegraphics[width=\textwidth]{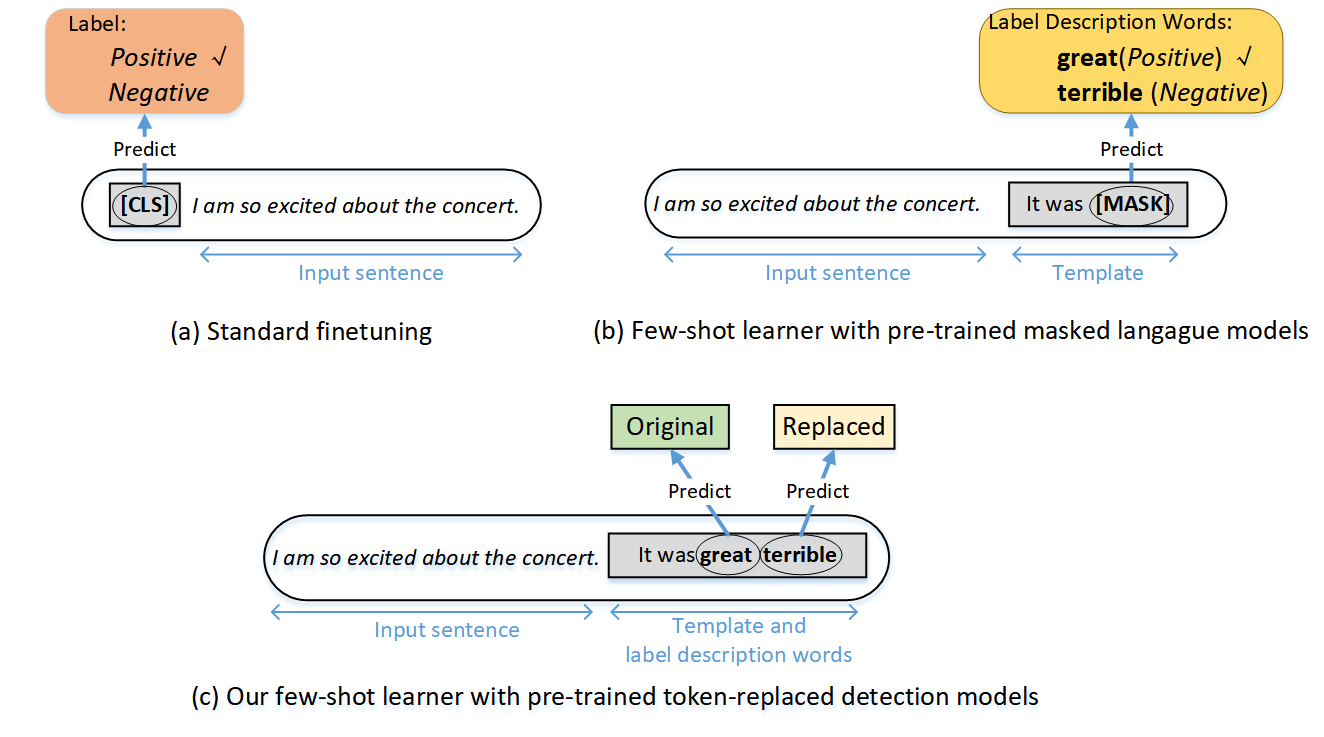}
    \caption{Different approaches of applying pre-trained models to sentiment classification.}
    \label{fig:fig1}
  \end{figure*}

  Some large pre-trained language models such as GPT-3 \citep{brown2020language} have achieved remarkable few-shot performance by reformulating tasks as language model problems. However, its hundreds of billions of parameters deter researchers and practitioners from applying it widely. To tackle this, a new paradigm, equipping smaller masked language models \citep{devlin2018bert} with few-shot capabilities \citep{schick2020exploiting,schick2020s,gao2020making} has been explored, wherein downstream tasks are treated as cloze questions. Typically, as illustrated in Figure \ref{fig:fig1}(b), each input sentence is appended with a prompt phrase such as ``It was [MASK]'' to each input sentence, allowing the model to fill in the [MASK] by reusing the masked language model head.
  
  Instead of masked language models, another self-supervised pre-training task called token-replaced detection has been proposed by \citet{clark2020electra} and it trains a model named ELECTRA to distinguish whether each token is replaced by a generated sample or not. One major advantage of token-replaced detection pre-training modeling is that it is more computationally efficient than masked language modeling. Moreover, their research demonstrates that given the same model size, pre-trained token-replaced detection models achieve substantially better performance than the pre-trained masked language  model such as BERT \cite{devlin2018bert} and RoBERTa \cite{liu2019roberta} in many downstream tasks.
  
  In this paper, inspired by the above unique effectiveness of the pre-trained token-replaced detection model, we propose a new approach, pre-trained token-replaced detection models as few-shot learners, aiming to further improve the few-shot learning performances. The key idea of our approach is to reformulate downstream tasks as token-replaced detection problems. Specifically, we first define a template and label description words which will be used to convert the input sentence into a prompted text. Then, we directly insert the template and all label description words into the sentence to form a prompt that might be an ungrammatical sentence. The motivation of this operation is to make our inputs similar to those in the data for training ELECTRA, having some replaced tokens. Lastly, we use a pre-trained token-replaced detection model to distinguish which label description word is the most original (i.e., least replaced) among all label description words. For instance, as illustrated in \ref{fig:fig1}(c), when performing a sentiment classification task, the input sentence ``\emph{I am so excited about the concert.}'' is converted into a new one ``\emph{I am so excited about the concert.} It was \textbf{great terrible}'' where ``\textbf{great}'' and ``\textbf{terrible}'' are two label description words for the two sentimental categories: \emph{positive} and \emph{negative}. Consequently, the pre-trained token-replaced detection model may predict the label description word ``\textbf{great}'' is more original (i.e., less replaced) than ``\textbf{terrible}'', which indicates that the input belongs to a \emph{positive} category. Compared to few-shot learners with pre-trained masked language models, in general, there are two major differences as follows. First, the designed phrases of few-shot learners with pre-trained masked don't contain any label description word. However, in our approach, we put them in prompts directly, which is easier to understand. Second, few-shot learners with pre-trained masked language models predict which label description word is the most appropriate to fill in [MASK], but our approach predicts which label description word is the most original (i.e., least replaced).
  
  To evaluate the few-shot capacity of our approach, we use both ELECTRA-Base and ELECTRA-Large as pre-trained token-replaced detection models to perform few-shot learning in our approach and conduct experiments in a wide variety of both one-sentence and two-sentence tasks. Empirical studies demonstrate that our approach outperforms few-shot learners with pre-trained masked language models.
  
  The contributions of this study are as follows:
  \begin{itemize}
    \item We propose a new approach for few-shot learning, which is simple and effective. To the best of our knowledge, few-shot learners with pre-trained token-replaced detection models is a novel branch of research that has not been explored in few-shot learning studies.
    \item A systematic evaluation of 16 popular datasets demonstrates that when given only a small number of labeled samples per class, our approach outperforms few-shot learners with pre-trained masked language models on most of these tasks.
  \end{itemize}

  The remainder of this paper is organized as follows. Section \ref{related work} overviews related studies about few-shot learning approaches and pre-trained token-replaced detection models. Section \ref{our approach} proposes our few-shot learner with a pre-trained token-replaced detection model in detail. Section \ref{experiments} presents the experimental results and analysis. Finally, Section \ref{conclusion} discusses the conclusions and future work.

  \section{Related Work}
  \label{related work}
  \subsection{Prompt-based few-shot learning}

  Few-shot learning with language model prompting has arisen with the introduction of GPT-3 \citep{brown2020language}, which adds a task description (prompt) with a training example demonstration to make the language model a few-shot learner. GPT-3's naive ``in-context learning'' paradigms have been applied to various tasks such as text classification \citep{min2021noisy,lu2021fantastically}, question answering \citep{liu2021makes}, and information extraction \citep{zhao2021calibrate}, which shows that a large pre-trained language model can achieve remarkable performance with only a few annotated samples. However, GPT-3's dependence on gigantic pre-trained language models narrows its scope of real applications.
  
  Instead of using a gigantic pre-trained language model, \citet{schick2020exploiting,schick2020s} reformulates a natural language processing (NLP) task as a cloze-style question with smaller masked language models \citep{devlin2018bert}. Their results show that it is possible to achieve few-shot performance similar to GPT-3 with much smaller language models. Due to the instability of manually designed prompts, many subsequent studies explore automatically searching the prompts, either in a discrete space \citep{gao2020making,jiang2020can,haviv2021bertese,shin2020autoprompt,ben2021pada} or in a continuous space \citep{qin2021learning,hambardzumyan2021warp,han2021ptr,liu2021gpt,zhang2022differentiable}. The discrete prompt is usually designed as natural language phrases with blank to be filled while the continuous prompt is a sequence of vectors that can be updated arbitrarily during learning. For instance, LM-BFF \citep{gao2020making} employs pre-trained mask language models and generates discrete prompts automatically. \citet{liu2021gpt} propose a prompt-based approach named P-tuning, which searches prompts in the continuous space by LSTM. \citet{zhang2022differentiable} propose another prompt-based approach named DifferentiAble pRompT (DART), which optimizes the prompt templates and the target labels differentially.
  
  
  Different from all existing above few-shot learning approaches, our approach reformulates NLP tasks as token-replaced detection problems and leverages label description words in the prompt.
  
  \subsection{Token-replaced detection}
  The token-replaced detection pre-training task is first introduced by \citet{clark2020electra}. Similar to the structure of GAN \citep{goodfellow2014generative}, it pre-trains a small generator to replace some tokens in an input with their plausible alternatives and then a large discriminator to distinguish whether each word has been replaced by the generator or not. The unique effectiveness of the pre-trained token-replaced detection model intrigues many studies to apply it in many NLP tasks, such as fact verification \citep{naseer2021empirical}, question answering \citep{alrowili2021large,yamada2021efficient}, grammatical error detection \cite{yuan2021multi}, emotional classification \citep{zhang2021emotional,guven2021effect}, and medication mention detection \citep{lee2020medication}. There are also some other studies that upgrade or extend the token-replaced detection pre-training mechanism. For instance, \citet{meng2021pretraining} jointly train multiple generators of different sizes to provide training signals at various levels of difficulty. \citet{futami2021asr} transfer the mechanism to visual pre-training and \citet{fang2022corrupted} propose an extended version of ELECTRA for speech recognition. 
  
  Different from all the above studies, to the best of our knowledge, this paper is the first study to apply pre-trained token-replaced detection models to few-shot learning.

  \section{Our approach}
  \label{our approach}
  A pre-trained token-replaced detection model like ELECTRA \citep{clark2020electra} trains a discriminator $\mathcal{D}$ that detects whether a token $x_{t}$ in an input token sequence $x=\left[x_{1}, \ldots, x_{t}, \ldots, x_{n}\right]$ is an original or replaced one. Suppose that the output from the discriminator is $y=\left[y_{1}, \ldots, y_{t}, \ldots, y_{n}\right]$. Then, $y_{t}=0$ (or 1) indicates that $x_{t}$ (at the position $t$) is an original (or a replaced) token. Specifically, in ELECTRA \citep{clark2020electra}, the discriminator is trained together with a masked language modeling generator which is used to generate replaced tokens in a token sequence. Finally, the discriminator performs the prediction with a sigmoid output layer, i.e.,
  \begin{equation}
    P\left(y_{t} \mid x_{t}\right)=\operatorname{sigmoid}\left(w^{T} h_{D}\left(x_{t}\right)\right)
  \end{equation}
  where $h_{D}\left(x_{t}\right)$ is the encoder function in the discriminator $\mathcal{D}$.

  \subsection{Few-shot Classification}
  \subsubsection{Few-shot Fine-tuning Phase}
  
  Suppose that the downstream classification task is a one-sentence classification problem and it has $k$ labels with label space $Y$ where $|Y|=k$. For the $i$-th category, we hand-craft a label description word $
  \mathbf { LABEL(i) }
  $. Then an input $x$ is rewritten as a prompt as follows:
  \begin{equation}
    \begin{aligned}
  x_{\text {prompt }}=&x \text { It was } \mathbf{L A B E L}(\mathbf{1})\ldots \\ &\mathbf{L A B E L}(\mathbf{i}) \ldots \mathbf{L A B E L}(\mathbf{k})
    \end{aligned}
  \end{equation}
  When the downstream classification task is a two-sentence classification problem, the input $\left(x_{1}, x_{2}\right)$ is rewritten as a prompt as follows:
  \begin{equation}
  \begin{aligned}
    x_{\text {prompt }}=&<x_{1}>\text{? } \mathbf{L A B E L}(\mathbf{1}) \ldots \\ & \mathbf{L A B E L}(\mathbf{i}) \ldots \mathbf{L A B E L}(\mathbf{k})\text{, }<x_{2}>
  \end{aligned}
\end{equation}
  
  Suppose that this sample belongs to the $i$-th label and the positions of the label description words are $\left[t_{1}, \ldots, t_{i}, \ldots, t_{k}\right]$. Thus, the output of the $i$-th description word $y_{t_{i}}$ is set to be 0 and this label description word is considered as original. In contrast, the outputs of all the other label description words are set to be 1 and these description words are considered as replaced. Note that the outputs of all tokens beyond label description words are set to be 0. Formally, the output of the whole prompt is obtained as follows:
  \begin{equation}
  \begin{aligned}
    y_{\text {prompt }}=&\left[\ldots, y_{t_{1}-1}=0, y_{t_{1}}=1, \ldots, y_{t_{i}}=0, \right. \\ & \left. \ldots, y_{t_{k}}=1, y_{t_{k}+1}=0, \ldots\right]
  \end{aligned}
  \end{equation}
  
  For instance, in a 5-category sentiment classification task, an input $x=$"\emph{This is one of his best films.}" could be rewritten as $x_{\text {prompt }}=$ "\emph{This is one of his best films.} It was \textbf{great good okay bad terrible}" where "\textbf{great}", "\textbf{good}", "\textbf{okay}", "\textbf{bad}", and "\textbf{terrible}" are used as the label description words for the \emph{very positive}, \emph{positive}, \emph{neutral}, \emph{negative}, and \emph{very negative} category. The output of  $x_{\text {prompt }}$ becomes $y_{\text {prompt }}=[\ldots,0,0,1,1,1,1]$.
  
  All prompt samples, together with their labels are used to update the parameters in the discriminator $\mathcal{D}$ of the pre-trained token-replaced detection model. Specifically, following the original progress of pre-training a token replaced model, we train the discriminator $\mathcal{D}$ by minimizing the binary cross entropy loss. It is important to note that our approach reuses the pre-trained weights ${w}^{T}$ in the formula (1) and does not use any other new parameters.
  
  \begin{table*}
    \centering
    \begingroup
    
    \begin{tabular}{llllll}
    \hline
    Category                                          & Dataset & $|Y|$ & \#Train & \#Test & Type                       \\
    \hline
    \multicolumn{1}{c}{\multirow{8}{*}{One-sentence}} & SST-2   & 2          & 6,920   & 872    & sentiment                  \\
    \multicolumn{1}{c}{}                              & SST-5   & 5          & 8,544   & 2,210   & sentiment                  \\
    \multicolumn{1}{c}{}                              & MR      & 2          & 8,662   & 2,000   & sentiment                  \\
    \multicolumn{1}{c}{}                              & CR      & 2          & 1,775   & 2,000   & sentiment                  \\
    \multicolumn{1}{c}{}                              & MPQA    & 2          & 8,606   & 2,000   & opinion polarity           \\
    \multicolumn{1}{c}{}                              & Subj    & 2          & 8,000   & 2.000   & subjectivity               \\
    \multicolumn{1}{c}{}                              & TREC    & 6          & 5,452   & 500    & question classification    \\
    \multicolumn{1}{c}{}                              & CoLA    & 2          & 8,551   & 1,042   & acceptability              \\
    \hline
    \multirow{8}{*}{Two-sentence}                     & MNLI    & 3          & 392,702 & 9,815  & natural language inference \\
                                                      & MNLI-MM & 3          & 392,702 & 9,832   & natural language inference \\
                                                      & SNLI    & 3          & 549,367 & 9,842  & natural language inference \\
                                                      & QNLI    & 2          & 104,743 & 5,463  & natural language inference \\
                                                      & RTE     & 2          & 2,490   & 277    & natural language inference \\
                                                      & MRPC    & 2          & 3,668   & 408    & paraphrase                 \\
                                                      & QQP     & 2          & 363,846 & 40,431 & paraphrase                 \\
                                                      & STS-B   & $R$ & 5,749   & 1,500  & sentence similarity       \\ \hline
    \end{tabular}
    \endgroup
    \caption{The details of 16 datasets: $|Y|$: \# of classes for classification tasks (Note that STS-B is a regression task over a bounded interval {[}0, 5{]}). In our few-shot experiments, we train and develop on limited examples sampled from the original training set and evaluate on the complete test set.}
    \label{datasets}
     \end{table*}

  \subsubsection{Testing Phase}
  In the testing phase, a testing sample is rewritten as a prompt according to formula (2) or (3) and the labels of all label description words in this prompt is predicted with the following formula, i.e.,
  \begin{equation}
    \begin{aligned}
    P(y \mid \mathbf{L A B E L}(\mathbf{i}))=& \operatorname{sigmoid} \left( \right.\\ & \left. w^{T} h_{D}(\mathbf{L A B E L}(\mathbf{i}))\right)
  \end{aligned}
  \end{equation}
  Then, the real label of the sample, i.e., $l_{\text {test }}$, is determined by the following formula, i.e.,
  \begin{equation}
    l_{\text {test }}=\underset{i}{\operatorname{argmax}}(P(y=0 \mid \mathbf{L A B E L}(\mathbf{i})))
  \end{equation}
  
  \subsection{Few-shot Regression}
  
  \subsubsection{Few-shot Fine-tuning Phase}
  Suppose that the downstream task is a regression problem and it has label space $Y$ where $Y$ is a bounded interval $\left[v_{l}, v_{u}\right]$. Following \citet{gao2020making}, we reformulate the problem as a "binary classification"---predicting the probabilities of belonging to two opposing poles, $\left\{c_{l}, c_{u}\right\}$ with values $v_{l}$ and $v_{u}$ respectively. 
  
  Then, a few-shot regression problem can be handled as a few-shot classification problem that has two labels with label space $\left\{c_{l}, c_{u}\right\}$. Same as classification tasks above, we rewrite an input $x$ as a prompt for a one-sentence regression task as follows:
  \begin{equation}
  x_{\text {prompt }}=x \text { It was } \mathbf{LABEL}(l) \mathbf{L A B E L}(u)
  \end{equation}
  When the downstream task is a two-sentence regression task, we rewrite an input $x$ as a prompt as follows:
  \begin{equation}
    \begin{aligned}
  x_{\text {prompt }}=&<x_{1}>\mathbf{L A B E L}(l) \mathbf{L A B E L}(u)\text{,} \\ & <x_{2}>
    \end{aligned}
  \end{equation}
  where $\mathbf { LABEL }(l)$ and $\mathbf  { LABEL }(u)$ denote the label description words for the low and upper bound categories.
  
  Suppose that the positions of the two label description words are $\left[t_{l}, t_{u}\right]$. Then, the output of the whole prompt is obtained as follows:
  \begin{equation}
    \begin{aligned}
  y_{\text {prompt }}=&\left[\ldots, y_{t_{l}-1}=0, y_{t_{l}}=\left(1-P\left(c_{l} \mid x\right)\right), \right.\\ &\left. y_{t_{u}}=\left(1-P\left(c_{u} \mid x\right)\right), y_{t_{u}+1}=0, \ldots\right]
    \end{aligned}
  \end{equation}
  where  $P\left(c_{l} \mid x\right)$  and  $P\left(c_{u} \mid x\right)$ are the posterior probabilities of $x$ belonging to $c_{l}$ and $c_{u}$ and satisfy the equation, i.e., $P\left(c_{l} \mid x\right)+P\left(c_{u} \mid x\right)=1 $. Following \citet{gao2020making}, these two probabilities could be estimated as follows:
  \begin{align}
  P\left(c_{l} \mid x\right)=\frac{v_{u}-y}{v_{u}-v_{l}} \\
  P\left(c_{u} \mid x\right)=\frac{y-v_{l}}{v_{u}-v_{l}}
  \end{align}
 
  For instance, in a two-sentence similarity regression task over the interval [0, 5], the label value of two sentences "\emph{Kittens are eating food.}" and "\emph{Kittens are eating from dishes.}" is 4.0. We use \textbf{No} and \textbf{Yes} as the label description words for the low and upper bound categories. According to formula (8-11), we construct its $x_{\text {prompt }}$ as "\emph{Kittens are eating food.} \textbf{No Yes}, \emph{Kittens are eating from dishes.}" and obtain its output $y_{\text {prompt }}=[\ldots, 0, \mathbf{0 . 8}, \mathbf{0 . 2}, 0, \ldots]$. 
  
  Same as classification tasks, we also adopt binary cross entropy loss and utilize all prompt samples together with their labels to fine-tune the discriminator $\mathcal{D}$. It is important to note that our approach reuses the pre-trained weights ${w}^{T}$ in the formula (1) and does not use any other new parameters.
  
  \subsubsection{Testing Phase}
  
  In the testing phase, a testing sample $x_{\text {test}}$ is rewritten as a prompt according to formula (7) or (8) and the outputs of the few-shot learner is obtained with the following formula, i.e.,
    \begin{align}
  y_{l}&=\operatorname{sigmoid}\left(w^{T} h_{D}(\mathbf{L A B E L}(l))\right) \\ 
  y_{u}&=\operatorname{sigmoid}\left(w^{T} h_{D}(\mathbf{L A B E L}(u))\right)
    \end{align}
  From formula (9), we can get
  \begin{align}
  P\left(c_{l} \mid x_{\text {test }}\right)&=1-y_{l} \\
  P\left(c_{u} \mid x_{\text {test }}\right)&=1-y_{u}
  \end{align}  
  Note that these two posterior probabilities might not satisfy the equation, i.e., $P\left(c_{l} \mid x\right)+P\left(c_{u} \mid x\right)=1 $. Therefore, we use a normalization method to update the two probabilities, i.e.,
    \begin{align}
    P^{\prime}\left(c_{l} \mid x_{\text {test }}\right)=\frac{P\left(c_{l} \mid x_{\text {test }}\right)}{\left(\left(P\left(c_{l} \mid x_{\text {test }}\right) +P\left(c_{u} \mid x_{\text {test }}\right)\right)\right.} \\
    P^{\prime}\left(c_{u} \mid x_{\text {test }}\right)=\frac{P\left(c_{u} \mid x_{\text {test }}\right)}{\left(\left(P\left(c_{l} \mid x_{\text {test }}\right)+P\left(c_{u} \mid x_{\text {test }}\right)\right)\right.}
  \end{align}
  Then, the regression value of the test sample, i.e., $v_{\text {test }}$, is obtained by using the following formula \citep{gao2020making}:
  \begin{equation}
  v_{\text {test }}=v_{l} \cdot P^{\prime}\left(\mathrm{c}_{l} \mid x_{\text {test }}\right)+v_{u} \cdot P^{\prime}\left(c_{u} \mid x_{\text {test }}\right)
  \end{equation}

    \begin{table*}
      \begingroup
      \setlength{\tabcolsep}{6pt} 
      \renewcommand{\arraystretch}{1} 
        \centering
        \resizebox{\textwidth}{!}{
      \begin{tabular}{llll}
        \hline
      Task &
        Template &
        Label Space &
        \textbf{Label(1) \ldots Label(k)} \\
        \hline
      \multicolumn{4}{l}{\textbf{One-sentence}} \\
      SST-2 &
        \emph{\textless{}S1\textgreater} It was  \textbf{Label(1) \ldots Label(k)} &
        \textit{positive, negative} &
        \textbf{great, terrible} \\
      SST-5 &
        \emph{\textless{}S1\textgreater} It was  \textbf{Label(1) \ldots Label(k)} &
        \begin{tabular}[c]{@{}l@{}} \textit{very positive, positive, neutral,} \\ \quad \textit{negative, very negative} \end{tabular} &
        \begin{tabular}[c]{@{}l@{}} \textbf{great, good, okay, }\\ \quad \textbf{bad, terrible} \end{tabular} \\
      MR &
        \emph{\textless{}S1\textgreater} It was  \textbf{Label(1) \ldots Label(k)} &
        \textit{positive, negative} &
        \textbf{great, terrible} \\
      CR &
        \emph{\textless{}S1\textgreater} It was  \textbf{Label(1) \ldots Label(k)} &
        \textit{positive, negative} &
        \textbf{great, terrible} \\
      MPQA &
        \emph{\textless{}S1\textgreater} It was  \textbf{Label(1) \ldots Label(k)} &
        \textit{positive, negative} &
        \textbf{great, terrible} \\ 
      Subj &
        \emph{\textless{}S1\textgreater} This is  \textbf{Label(1) \ldots Label(k)} &
        \textit{subjective, objective} &
        \textbf{subjective, objective} \\
      TREC &
        \textbf{Label(1) \ldots Label(k)}: \emph{\textless{}S1\textgreater}{} & 
        \begin{tabular}[c]{@{}l@{}} \textit{abbreviation, entity, description,} \\ \textit \quad \textit{human, location, numeric} \end{tabular} &
         \begin{tabular}[c]{@{}l@{}} \textbf{Expression, Entity, Description,} \\ \quad \textbf{Human, Location, Number} \end{tabular} \\
      COLA &
        \emph{\textless{}S1\textgreater} This is  \textbf{Label(1) \ldots Label(k)} &
        \textit{grammatical, not\_grammatical} &
        \textbf{correct, incorrect} \\

        \hline
      \multicolumn{4}{l}{\textbf{Two-sentence}} \\
      MNLI &
        \emph{\textless{}S1\textgreater} ? \textbf{Label(1) \ldots Label(k)}, \emph{\emph{\textless{}S2\textgreater}}{} &
        \textit{entailment, neutral, contradiction} &
        \textbf{Yes, Maybe, No} \\
      MNLI-MM &
        \emph{\textless{}S1\textgreater} ? \textbf{Label(1) \ldots Label(k)}, \emph{\emph{\textless{}S2\textgreater}}{} &
        \textit{entailment, neutral, contradiction} &
        \textbf{Yes, Maybe, No} \\
      SNLI &
        \emph{\textless{}S1\textgreater} ? \textbf{Label(1) \ldots Label(k)}, \emph{\emph{\textless{}S2\textgreater}}{} &
        \textit{entailment, neutral, contradiction} &
        \textbf{Yes, Maybe, No} \\
      QNLI &
        \emph{\textless{}S1\textgreater} ? \textbf{Label(1) \ldots Label(k)}, \emph{\emph{\textless{}S2\textgreater}}{} &
        \textit{entailment, not\_entailment} &
        \textbf{Yes, No} \\
      RTE &
        \emph{\textless{}S1\textgreater} ? \textbf{Label(1) \ldots Label(k)}, \emph{\emph{\textless{}S2\textgreater}}{} &
        \textit{entailment, not\_entailment} &
        \textbf{Yes, No} \\
      MRPC &
        \emph{\textless{}S1\textgreater} \textbf{Label(1) \ldots Label(k)}, \emph{\emph{\textless{}S2\textgreater}}{} &
        \textit{equivalent, not\_ equivalent} &
        \textbf{Yes, No} \\
      QQP &
        \emph{\textless{}S1\textgreater} \textbf{Label(1) \ldots Label(k)}, \emph{\emph{\textless{}S2\textgreater}}{} &
        \textit{equivalent, not\_ equivalent} &
        \textbf{Yes, No} \\
      STS-B &
        \emph{\textless{}S1\textgreater} \textbf{Label(1) \ldots Label(k)}, \emph{\emph{\textless{}S2\textgreater}}{} &
        \textit{{[}0,5{]}} &
        \textbf{Yes, No} \\ \hline
      \end{tabular}}
      \endgroup
      \caption{Manual templates and label description words in our experiments.}
      \label{template}
       \end{table*}
  
  \section{Experiments}
  \label{experiments}
  In this section, we compare our approach with a few-shot learning approach based on pre-trained masked language models. Furthermore, we evaluate the impact of different templates, label description words and training data scales.
  
  \subsection{Evaluation Setting}
  We conduct a systematic empirical study based on the datasets used in \citet{gao2020making}. The experimental data contains 16 datasets from many kinds of NLP tasks such as sentiment analysis, question classification, opinion polarity, subjectivity, acceptability, natural language inference, paraphrase, and sentence similarity \citep{wang2018glue,bowman2015large}. Following \citet{gao2020making}, we divide these tasks into two categories, i.e., one-sentence (single sentence) input and two-sentence (sentence pair) input tasks. In addition, these tasks not only contain binary or multi-class classification but also contain regression. See the statistics of datasets in Table \ref{datasets}.
  
  \subsection{Evaluation protocol}
  Note that the results of few-shot learning experiments are very sensitive and unstable to the different splits of data and hyper-parameter setups \cite{dodge2020fine,zhang2020revisiting}, because the size of the training examples is so small. Thus, we follow the evaluation protocol of \cite{gao2020making} by running 5 experiments with 5 different training and development splits, randomly sampled from the original training set using a fixed set of seeds, and then measuring the average results and standard deviations. Note that, following \cite{gao2020making}, we sample the same size of development set as the training set. For the hyper-parameters, we also utilize grid search to get the best hyper-parameter setup. We set the weight\_decay to be 2e-3, max\_length to be 256 and use AdamW optimizer with epsilon 1e-8. We change the learning rate in the set of $\{$1e-5,2e-5,3e-5,4e-5,5e-5$\}$ and the batch size between 4 or 8. Besides, we use manual templates and label description words for each task and the details are shown in Table \ref{template}.
  
  \begin{table*}[h]
    \begingroup
    \renewcommand{\arraystretch}{1} 
      \centering
      \resizebox{\textwidth}{!}{
        \begin{tabular}{llllllllll}
          \hline
          \textbf{One-sentence} &
            \multicolumn{1}{c}{\begin{tabular}[c]{@{}c@{}}SST-2\\ (acc)\end{tabular}} &
            \multicolumn{1}{c}{\begin{tabular}[c]{@{}c@{}}SST-5\\ (acc)\end{tabular}} &
            \multicolumn{1}{c}{\begin{tabular}[c]{@{}c@{}}MR\\ (acc)\end{tabular}} &
            \multicolumn{1}{c}{\begin{tabular}[c]{@{}c@{}}CR\\ (acc)\end{tabular}} &
            \multicolumn{1}{c}{\begin{tabular}[c]{@{}c@{}}MPQA\\ (acc)\end{tabular}} &
            \multicolumn{1}{c}{\begin{tabular}[c]{@{}c@{}}Subj\\ (acc)\end{tabular}} &
            \multicolumn{1}{c}{\begin{tabular}[c]{@{}c@{}}TREC\\ (acc)\end{tabular}} &
            \multicolumn{1}{c}{\begin{tabular}[c]{@{}c@{}}CoLA\\ (matt)\end{tabular}} &
            \multicolumn{1}{c}{\textit{AVG}} \\ \hline
          \begin{tabular}[c]{@{}l@{}}Fine-tuning(RoBERTa)\end{tabular} &
            77.8 (2.8) &
            38.5 (1.6) &
            70.1 (4.9) &
            76.7 (2.8) &
            70.1 (8.0) &
            89.7 (0.8) &
            81.5 (4.3) &
            18.9 (11.7) &
            \textit{65.4} \\
          \begin{tabular}[c]{@{}l@{}}Fine-tuning(ELECTRA)\end{tabular} &
            82.8 (3.5) &
            41.6 (3.6) &
            73.9 (3.5) &
            82.9 (4.1) &
            70.7 (5.1) &
            \textbf{92.0 (0.5)} &
            78.5 (5.8) &
            \textbf{39.3 (3.4)} &
            \textit{70.2} \\
            \begin{tabular}[c]{@{}l@{}}P-tuning(RoBERTa)\end{tabular} &
              83.3 (5.3) &
              43.1 (2.1) &
              81.7 (1.2) &
              86.0 (3.6) &
              74.0 (5.2) &
              89.0 (1.1) &
              76.9 (8.3) &
              -0.8 (2.5) &
              \textit{66.7} \\
            
          \begin{tabular}[c]{@{}l@{}}LM-BFF(RoBERTa)\end{tabular} &
            87.2 (1.3) &
            44.5 (0.8) &
            83.4 (1.4) &
            89.1 (1.5) &
            81.3 (3.9) &
            89.3 (1.8) &
            77.5 (6.1) &
            5.3 (5.3) &
            \textit{69.7} \\
            \begin{tabular}[c]{@{}l@{}}DART(RoBERTa)\end{tabular} &
            88.9 (0.5) &
            45.3 (1.5) &
            83.7 (1.0) &
            89.2 (1.4) &
            76.6 (6.3) &
            88.9 (2.2) &
            77.3 (7.2) &
            4.2 (5.0) &
            \textit{69.3} \\
          \begin{tabular}[c]{@{}l@{}}Ours(ELECTRA)\end{tabular} &
            \textbf{91.7 (0.8)} &
            \textbf{49.7 (1.0)} &
            \textbf{86.8 (2.8)} &
            \textbf{90.8 (1.0)} &
            \textbf{84.5 (1.5)} &
            87.5 (1.2) &
            \textbf{82.2 (3.3)} &
            24.7 (11.8) &
            \textit{\textbf{74.7}} \\ \hline
          \textbf{Two-sentence} &
            \multicolumn{1}{c}{\begin{tabular}[c]{@{}c@{}}MNLI\\ (acc)\end{tabular}} &
            \multicolumn{1}{c}{\begin{tabular}[c]{@{}c@{}}MNLI-MM\\ (acc)\end{tabular}} &
            \multicolumn{1}{c}{\begin{tabular}[c]{@{}c@{}}SNLI\\ (acc)\end{tabular}} &
            \multicolumn{1}{c}{\begin{tabular}[c]{@{}c@{}}QNLI\\ (acc)\end{tabular}} &
            \multicolumn{1}{c}{\begin{tabular}[c]{@{}c@{}}RTE\\ (acc)\end{tabular}} &
            \multicolumn{1}{c}{\begin{tabular}[c]{@{}c@{}}MRPC\\ (f1)\end{tabular}} &
            \multicolumn{1}{c}{\begin{tabular}[c]{@{}c@{}}QQP\\ (f1)\end{tabular}} &
            \multicolumn{1}{c}{\textit{\begin{tabular}[c]{@{}c@{}}STS-B\\ (pear)\end{tabular}}} &
            \multicolumn{1}{c}{\textit{AVG}} \\ \hline
          \begin{tabular}[c]{@{}l@{}}Fine-tuning(RoBERTa)\end{tabular} &
            38.6 (2.5) &
            39.5 (2.7) &
            48.0 (4.7) &
            \textbf{63.2 (6.7)} &
            51.9 (1.6) &
            74.5 (4.4) &
            58.6 (6.0) &
            \textit{65.2 (8.7)} &
            \textit{54.9} \\
          \begin{tabular}[c]{@{}l@{}}Fine-tuning(ELECTRA)\end{tabular} &
            46.9 (3.6) &
            48.9 (3.8) &
            50.6 (2.1) &
            59.9 (2.3) &
            52.6 (2.5) &
            \textbf{76.9 (2.9)} &
            \textbf{64.1 (2.8)} &
            \textit{\textbf{72.4 (2.0)}} &
            \textit{59.0} \\
          \begin{tabular}[c]{@{}l@{}}P-tuning(RoBERTa)\end{tabular} &
            50.6 (1.1) &
            50.6 (1.1) &
            55.0 (4.3) &
            58.1 (3.1) &
            56.0 (4.2) &
            70.2 (2.3) &
            58.7 (2.8) &
            \textit{-} &
            \textit{57.0} \\

          \begin{tabular}[c]{@{}l@{}}LM-BFF(RoBERTa)\end{tabular} &
            59.1 (2.4) &
            60.9 (2.4) &
            64.3 (2.9) &
            61.8 (4.8) &
            57.9 (6.7) &
            72.3 (6.6) &
            62.7 (2.1) &
            \textit{68.6 (5.7)} &
            \textit{63.5} \\
          \begin{tabular}[c]{@{}l@{}}DART(RoBERTa)\end{tabular} &
            55.3 (2.4) &
            55.3 (2.4) &
            62.6 (2.6) &
            58.4 (4.3) &
            58.2 (6.0) &
            72.4 (2.5) &
            60.4 (1.7) &
            \textit{-} &
            \textit{60.4} \\
          \begin{tabular}[c]{@{}l@{}}Ours(ELECTRA)\end{tabular} &
            \textbf{59.7 (2.4)} &
            \textbf{61.8 (2.0)} &
            \textbf{68.9 (3.2)} &
            61.9 (2.4) &
            \textbf{61.5 (2.9)} &
            73.9 (3.9) &
            58.0 (3.8) &
            \textit{66.6 (2.9)} &
            \textit{\textbf{64.0}} \\ \hline
          \end{tabular}}
    \endgroup
    \caption{Experimental results of different approaches when base pre-trained models are used.}
    \label{base}
    \end{table*}
  \subsection{Main results}
  We use 16 samples per class for few-shot learning experiments and conduct our experiments on both base-level and large-level pre-trained model scenarios. We compare our approach with several baselines including 1) Fine-tuning: standard fine-tuning of pre-trained models; 2) P-tuning \citep{liu2021gpt}: few-shot learner that searches prompts in a continuous space by LSTM;  3) LM-BFF \citep{gao2020making}: few-shot learner that employs pre-trained mask language models and discrete prompts. For a fair comparison, we use the same templates and label description words as our approach and do not use any demonstrations; 4) DART \citep{zhang2022differentiable}: few-shot learner that optimizes the prompt templates and the target labels differentially.

  \subsubsection{RoBERTa-Base and ELECTRA-Base Results}
  Table \ref{base} gives the experimental results of different prompt-based approaches to few-shot learning with a base pre-trained model, i.e., RoBERTa-Base or ELECTRA-Base. The best performance in each task is bold in the table. Note that since there is no implementation for regressions tasks in the two baseline approaches, i.e., P-tuning and DART and thus we do not reproduce their approach on STS-B which is a regression task. From this table, we discuss the results in two scenarios, i.e., one-sentence and two-sentence tasks.
  
  In one-sentence tasks, first, fine-tuning with RoBERTa-Base performs worse than fine-tuning with ELECTRA-Base on average (65.4\% vs. 70.2\%), which indicates that ELECTRA-Base is a better fine-tuner even when only a few training samples are available. This result is consistent with the conclusion reported in \citet{clark2020electra} when many training samples are available.  Second, all prompt-based approaches greatly outperform standard fine-tuning on most tasks, which indicates that few-shot learners with either base masked language model or base token-replaced detection model are powerful in few-shot learning. One big exception is CoLA \cite{warstadt2019neural} where few-shot learning approaches perform much worse than fine-tuning approaches. This might be because the task aims to detect whether a sentence is grammatical or non-grammatical which is difficult to find suitable label description words. However, interestingly, we find that ELECTRA-Base performs much better than RoBERTa-Base in this task. Third, our approach yields excellent results and performs much better than P-tuning, LM-BFF and DART on average (74.7\% vs. 66.7\%, 69.7\% and 69.3\%), which encourages using a pre-trained token-replaced detection model for few-shot learning in one-sentence tasks.
  
  In two-sentence tasks, first, standard fine-tuning with RoBERTa-Base still performs worse than fine-tuning with ELECTRA-Base. Second, all prompt-based approaches greatly outperform standard fine-tuning on most tasks, which once again indicates that few-shot learners with either mask language model or token-replaced detection model are powerful in few-shot learning. Third, our approach performs better than P-tuning, LM-BFF and DART, although the average improvements are quite limited (64.0\% vs. 57\%, 63.5\% and 60.4\%).

  \begin{table*}
    \begingroup
  \renewcommand{\arraystretch}{1} 
    \centering
    \resizebox{\textwidth}{!}{
      \begin{tabular}{llllllllll}
        \hline
        \textbf{One-sentence} &
          \multicolumn{1}{c}{\begin{tabular}[c]{@{}c@{}}SST-2\\ (acc)\end{tabular}} &
          \multicolumn{1}{c}{\begin{tabular}[c]{@{}c@{}}SST-5\\ (acc)\end{tabular}} &
          \multicolumn{1}{c}{\begin{tabular}[c]{@{}c@{}}MR\\ (acc)\end{tabular}} &
          \multicolumn{1}{c}{\begin{tabular}[c]{@{}c@{}}CR\\ (acc)\end{tabular}} &
          \multicolumn{1}{c}{\begin{tabular}[c]{@{}c@{}}MPQA\\ (acc)\end{tabular}} &
          \multicolumn{1}{c}{\begin{tabular}[c]{@{}c@{}}Subj\\ (acc)\end{tabular}} &
          \multicolumn{1}{c}{\begin{tabular}[c]{@{}c@{}}TREC\\ (acc)\end{tabular}} &
          \multicolumn{1}{c}{\begin{tabular}[c]{@{}c@{}}CoLA\\ (matt)\end{tabular}} &
          \multicolumn{1}{c}{\textit{AVG}} \\ \hline
        \begin{tabular}[c]{@{}l@{}}Fine-tuning(RoBERTa)\end{tabular} &
          81.4 (3.8) &
          43.9 (2.0) &
          76.9 (5.9) &
          75.8 (3.2) &
          72.0 (3.8) &
          90.8 (1.8) &
          \textbf{88.8 (2.1)} &
          \textbf{33.9 (14.3)} &
          \textit{70.4} \\
        \begin{tabular}[c]{@{}l@{}}Fine-tuning(ELECTRA)\end{tabular} &
          79.9 (7.9) &
          41.2 (1.9) &
          73.0 (5.4) &
          75.0 (6.4) &
          65.3 (6.9) &
          \textbf{94.0 (1.0)} &
          82.8 (8.0) &
          33.4 (10.4) &
          \textit{68.1} \\
          \begin{tabular}[c]{@{}l@{}}P-tuning(RoBERTa)\end{tabular} &
            89.6 (2.6) &
            48.0 (1.3) &
            85.4 (1.9) &
            88.7 (2.6) &
            76.3 (3.3) &
            90.9 (1.5) &
            86.2 (3.4) &
            4.0 (5.3) &
            \textit{71.1} \\
        \begin{tabular}[c]{@{}l@{}}LM-BFF(RoBERTa)\end{tabular} &
          92.7 (0.9) &
          47.4 (2.5) &
          87.0 (1.2) &
          90.3 (1.0) &
          \textbf{84.7 (2.2)} &
          91.2 (1.1) &
          84.8 (5.1) &
          9.3 (7.3) &
          \textit{73.4} \\
        \begin{tabular}[c]{@{}l@{}}DART(RoBERTa)\end{tabular} &
          91.6 (1.0) &
          47.4 (3.3) &
          85.7 (3.0) &
          90.3 (0.8) &
          66.6 (6.4) &
          89.9 (1.7) &
          84.8 (4.6) &
          10.0 (8.4) &
          \textit{70.8} \\     
        \begin{tabular}[c]{@{}l@{}}Ours(ELECTRA)\end{tabular} &
          \textbf{92.8 (0.6)} &
          \textbf{50.7 (2.9)} &
          \textbf{89.4 (0.8)} &
          \textbf{90.5 (2.2)} &
          83.2 (1.4) &
          92.1 (0.7) &
          87.2 (3.8) &
          16.3 (15.1) &
          \textit{\textbf{75.3}} \\ \hline
        \textbf{Two-sentence} &
          \multicolumn{1}{c}{\begin{tabular}[c]{@{}c@{}}MNLI\\ (acc)\end{tabular}} &
          \multicolumn{1}{c}{\begin{tabular}[c]{@{}c@{}}MNLI-MM\\ (acc)\end{tabular}} &
          \multicolumn{1}{c}{\begin{tabular}[c]{@{}c@{}}SNLI\\ (acc)\end{tabular}} &
          \multicolumn{1}{c}{\begin{tabular}[c]{@{}c@{}}QNLI\\ (acc)\end{tabular}} &
          \multicolumn{1}{c}{\begin{tabular}[c]{@{}c@{}}RTE\\ (acc)\end{tabular}} &
          \multicolumn{1}{c}{\begin{tabular}[c]{@{}c@{}}MRPC\\ (f1)\end{tabular}} &
          \multicolumn{1}{c}{\begin{tabular}[c]{@{}c@{}}QQP\\ (f1)\end{tabular}} &
          \multicolumn{1}{c}{\textit{\begin{tabular}[c]{@{}c@{}}STS-B\\ (pear)\end{tabular}}} &
          \multicolumn{1}{c}{\textit{AVG}} \\ \hline
        \begin{tabular}[c]{@{}l@{}}Fine-tuning(RoBERTa)\end{tabular} &
          45.8 (6.4) &
          47.8 (6.8) &
          48.4 (4.8) &
          60.2 (6.5) &
          54.4 (3.9) &
          76.6 (2.5) &
          60.7 (4.3) &
          \textit{53.5 (8.5)} &
          \textit{55.9} \\
        \begin{tabular}[c]{@{}l@{}}Fine-tuning(ELECTRA)\end{tabular} &
          54.4 (2.4) &
          56.7 (1.7) &
          58.8 (4.8) &
          62.9 (4.1) &
          53.8 (3.7) &
          \textbf{78.7 (3.1)} &
          67.2 (3.4) &
          \textit{\textbf{78.5 (0.5)}} &
          \textit{63.9} \\
          \begin{tabular}[c]{@{}l@{}}P-tuning(RoBERTa)\end{tabular} &
            59.7 (3.0) &
            59.7 (3.0) &
            71.8 (3.5) &
            62.5 (6.5) &
            61.8 (2.6) &
            72.7 (7.4) &
            64.2 (1.5) &
            \textit{-} &
            \textit{64.6} \\
          
        \begin{tabular}[c]{@{}l@{}}LM-BFF(RoBERTa)\end{tabular} &
          68.3 (2.3) &
          70.5 (1.9) &
          77.2 (3.7) &
          64.5 (4.2) &
          69.1 (3.6) &
          74.5 (5.3) &
          65.5 (5.3) &
          \textit{71.0(7.0)} &
          \textit{70.1} \\
        \begin{tabular}[c]{@{}l@{}}DART(RoBERTa)\end{tabular} &
          67.1 (2.6) &
          67.0 (2.5) &
          74.0 (4.0) &
          63.1 (3.0) &
          64.5 (5.2) &
          75.9 (4.7) &
          63.4 (4.4) &
          \textit{-} &
          \textit{67.9} \\
        \begin{tabular}[c]{@{}l@{}}Ours(ELECTRA)\end{tabular} &
          \textbf{69.2 (4.0)} &
          \textbf{71.0 (3.5)} &
          \textbf{79.3 (3.2)} &
          \textbf{69.0 (4.5)} &
          \textbf{74.2 (3.1)} &
          73.2 (7.5) &
          \textbf{68.2 (3.4)} &
          \textit{74.7 (2.9)} &
          \textit{\textbf{72.4}} \\ \hline
        \end{tabular}}
    \endgroup
    \caption{Experimental results of different approaches when large pre-trained models are used.}
    \label{large}
       \end{table*}
  
  \subsubsection{RoBERTa-Large and ELECTRA-Large Results}
  Table \ref{large} gives the experimental results of different prompt-based approaches to few-shot learning with a large pre-trained model, i.e., RoBERTa-Large or ELECTRA-Large. The best performance in each task is bold in the table. From this table, we discuss the results in two scenarios, i.e., one-sentence and two-sentence tasks.
  
  In one-sentence tasks, first, fine-tuning with RoBERTa-Large performs a bit better than fine-tuning with ELECTRA-Large on average (68.1\% vs. 70.4\%), which indicates that the choice of ELECTRA and RoBERTa might depend on the tasks when large models are used. Second, all prompt-based approaches greatly outperform standard fine-tuning on most tasks, which indicates that few-shot learners with either large masked language models or large token-replaced detection models are powerful in few-shot learning. However, CoLA is still the exception and even worse, the performance of few-shot learning with ELECTRA-Large performs worse than ELECTRA-Base, (16.3\% vs. 24.7\%). This result shows that the prompting style in our few-shot learning approach seems not suitable for the task of grammatical or non-grammatical detection. Third, our approach yields performances better than P-Tuning, LM-BFF and DART, achieving 4.2\%, 1.9\% and 4.5\% average improvements respectively.
  
  In two-sentence tasks, first, fine-tuning with RoBERTa-Large performs much worse than fine-tuning with ELECTRA-Large (55.9\% vs. 63.9\%). Second, all prompt-based approaches greatly outperform standard fine-tuning on many tasks, which once again indicates that few-shot learners with either mask language model or pre-trained token-replaced detection model are powerful in few-shot learning. Third, our approach performs better than P-Tuning, LM-BFF and DART on average (72.4\% vs. 64.6\%, 70.1\% and 67.9\%).
  

  \begin{table*}
    \begingroup
    \setlength{\tabcolsep}{6pt} 
    \renewcommand{\arraystretch}{1} 
      \centering
      \resizebox{\textwidth}{!}{
    \begin{tabular}{l|l|l|l|l}
    \hline
    Task &
      Template &
      \textbf{Label(1) \ldots Label(k)} &
      \begin{tabular}[c]{@{}l@{}}LM-BFF \qquad\qquad \\ (acc)\end{tabular} &
      \begin{tabular}[c]{@{}l@{}}Our approach\\ (acc)\end{tabular} \\ \hline
    \multirow{8}{*}{\begin{tabular}[c]{@{}l@{}}SST-2 \\ (\emph{positive/negative})\end{tabular}} &
      \multirow{4}{*}{\emph{\textless{}S1\textgreater} It was  \textbf{Label(1) \ldots Label(k)}} &
      \textbf{great, terrible} &
      88.6 (1.3) &
      91.4 (1.6) \\ \cline{3-5} 
     &                                                                                               & \textbf{good, bad}       & 88.9 (0.6) & 91.0 (2.0) \\ \cline{3-5} 
     &                                                                                               & \textbf{dog, cat}        & 85.2 (2.0) & 79.6 (7.3) \\ \cline{3-5} 
     &                                                                                               & \textbf{terrible, great} & 82.4 (3.3) & 89.2 (1.9) \\ \cline{2-5} 
     & \multirow{4}{*}{\textbf{\textbf{Label(1) \ldots Label(k)}} : \emph{\textless{}S1\textgreater}{}}                   & \textbf{great, terrible} & 85.6 (3.0) & 91.1 (1.2) \\ \cline{3-5} 
     &                                                                                               & \textbf{good, bad}       & 87.5 (0.4) & 90.8 (0.7) \\ \cline{3-5} 
     &                                                                                               & \textbf{dog, cat}        & 80.1 (3.5) & 69.7 (8.2) \\ \cline{3-5} 
     &                                                                                               & \textbf{terrible, great} & 67.4 (3.5) & 76.4 (9.6) \\ \hline
    \multirow{4}{*}{\begin{tabular}[c]{@{}l@{}}MNLI \\ (\emph{entailment/neutral} \\ \emph{\quad /contradiction}) \end{tabular}} &
      \emph{\textless{}S1\textgreater}? \textbf{Label(1) \ldots Label(k)}, \emph{\textless{}S2\textgreater}{} &
      \multirow{4}{*}{\textbf{Yes, Maybe, No}} &
      58.3 (2.4) &
      58.8 (2.5) \\ \cline{2-2} \cline{4-5} 
     & \emph{\textless{}S2\textgreater}. \textbf{Label(1) \ldots Label(k)}, \emph{\textless{}S1\textgreater}{}                 &                          & 58.7 (1.3) & 57.6 (2.5) \\ \cline{2-2} \cline{4-5} 
     & \emph{\textless{}S1\textgreater} \textbf{Label(1) \ldots Label(k)} \emph{\textless{}S2\textgreater}{}                     &                          & 56.4 (1.8) & 53.6 (2.2) \\ \cline{2-2} \cline{4-5} 
     & \emph{\textless{}S1\textgreater}{}. \textbf{Label(1) \ldots Label(k)}, this is good, \emph{\textless{}S2\textgreater}{} &                          & 54.0 (2.4) & 55.8 (3.5) \\ \hline
    \end{tabular}}
    \endgroup
    \caption{The impact of different templates and label description words.}
    \label{diff}
     \end{table*}
  
     \begin{figure*}[h]
      \centering
      \includegraphics[width=\textwidth]{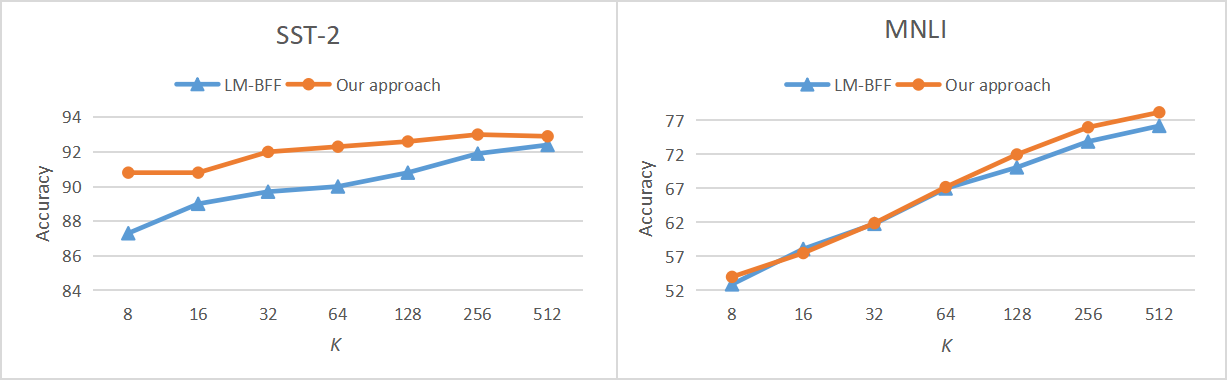}
      \caption{LM-BFF vs. our approach when using different numbers of labeled instances (\emph{K}: \# of labeled instances per class).}
      \label{fig:fig2}
      \end{figure*}

  \subsection{Impact of templates and label description words}
  We further conduct experiments on the one-sentence task SST-2 and the two-sentence task MNLI to study the impact of different templates and label description words in our approach. Due to a large number of trials in the grid search, we use a fixed batch size 4 and learning rate 2e-5 in this part. Table \ref{diff} shows the results of the LM-BFF approach with RoBERTa-Base, the best-performed approach in all prompt-based baselines, and our approach with ELECTRA-Base in the tasks of SST-2 and MNLI. From this table, we can see that the impact of different templates and label description words for our method is similar to LM-BFF. In terms of label description words, the more semantic-related the designed label words are to the categories, the more likely to achieve stable and excellent results. For instance, in SST-2, regardless of LM-BFF or our approach, the semantic-related label description words \textbf{great/terrible} and \textbf{good/bad} always outperform the words \textbf{dog/cat} and \textbf{terrible/great} which are semantically irrelevant or even opposite with the categories \emph{positive} and \emph{negative}. In terms of templates, the performance is a bit sensitive to the templates, even a punctuation mark. Besides, there seems to be no general principle to design templates to optimally adapt to our approach and LM-BFF. For instance, In MNLI, LM-BFF obtains the best performance with the template \emph{"<S1>.} \textbf{Label(1) \ldots Label(k),} \emph{ <S2>"}, while our approach obtains the best performance with the template \emph{"<S1>?} \textbf{Label(1) \ldots Label(k),} \emph{ <S2>"}. 
  
  \subsection{Impact of training data scales}
  We further conduct experiments on the one-sentence task SST-2 and the two-sentence task MNLI to study the impact of the numbers of labeled instances in our approach. In this part, we also use a fixed batch size 4 and learning rate 2e-5. Figure \ref{fig:fig2} shows the trends of the LM-BFF approach, the best-performed approach in all prompt-based baselines, and our approach when using different numbers of labeled instances. From this figure, we can see that our approach outperforms LM-BFF in different numbers of labeled instances in the one-sentence task SST-2. In the two-sentence task MNLI, our approach performs similarly to LM-BFF when the numbers of labeled instances are less than 64. But our approach outperforms LM-BFF when the numbers of labeled instances are among [128, 512].
  
  \section{Conclusion and Future Work}
  \label{conclusion}
  In this paper, we propose a novel few-shot learning approach with pre-trained token-replaced detection models, which transforms traditional classification and regression tasks into token-replaced detection problems. Empirical studies on 16 NLP datasets demonstrate that, in both one-sentence and two-sentence learning tasks, our approach generally achieves better performances in the few-shot scenario when compared to the masked language model-based few-shot learner. These results highlight that our approach is a comprehensive alternative for few-shot learning.

  In the future, we would like to explore the following directions. First, we notice that in some tasks like CoLA, standard fine-tuning is also a strong baseline and even performs much better than few-shot learners based on either a masked language model or a token-replaced detection model. Thus, it is interesting to combine [CLS] output vector, i.e., the standard fine-tuning style, with the prompting style, to further improve the few-shot learning performance. Second, we would like to apply our approach to some other NLP tasks, such as multi-label text classification and sequence labeling tasks like named entity recognition.

  \section*{Acknowledgement}
  This work was supported by a NSFC grant (No.62076176). We also acknowledge reviewers for their valuable suggestions.

\bibliography{compling_style}



\end{document}